\documentclass{article}

\newif\ifanon
\anonfalse

\ifanon
  \usepackage[dblblindworkshop]{neurips_2026}
\else
  \usepackage[final,sglblindworkshop]{neurips_2026}
\fi
\workshoptitle{XAI4Science: Knowledge Discovery and Trust through Interpretable Foundation Models}

\usepackage[utf8]{inputenc}
\usepackage[T1]{fontenc}
\usepackage{hyperref}
\hypersetup{colorlinks=true,allcolors=[rgb]{0.10,0.25,0.55},
            pdftitle={Candidates That Carry Nothing Pass the Acceptance Tests of Automated
                      Discovery and Interpretability},
            pdfauthor={}, pdfsubject={}, pdfkeywords={}, pdfcreator={}, pdfproducer={}}
\usepackage{url}
\usepackage{booktabs}
\usepackage{array}
\newcolumntype{L}[1]{>{\raggedright\arraybackslash}p{#1}}   
\usepackage{amsmath}
\usepackage{amssymb}
\usepackage{graphicx}
\usepackage{microtype}
\usepackage{tikz}
\usetikzlibrary{arrows.meta,positioning,fit,backgrounds}

\title{Null Controls Are Necessary but Not Sufficient: Evaluating Acceptance Guards for Scientific Discovery and Interpretability}

\ifanon
  \author{}
\else
  \hypersetup{pdfauthor={Project Aether; Tianyu Wu}}
  \author{%
    \raisebox{-0.32\height}{\includegraphics[height=1.7em]{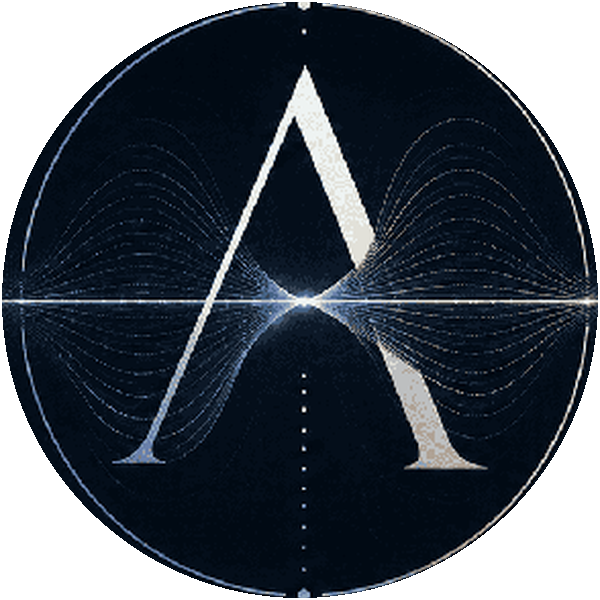}}\;%
    \textbf{Project Aether}\\[3pt]
    Tianyu Wu\quad\texttt{tianyu16@illinois.edu}\\
  }
\fi

\newcommand{\dystsN}{129}
\newcommand{\dystsConstCloses}{124}
\newcommand{\dystsConstPct}{96\%}

\newcommand{\dystsConstNonfinite}{5}

\newcommand{\mildN}{129}
\newcommand{\mildSlowCloses}{1}
\newcommand{\mildSlowBoth}{0}
\newcommand{\mildConstCloses}{124}
\newcommand{\mildConstPays}{0}
\newcommand{\mildEmptyBoth}{0}

\newcommand{\posRecallEst}{0.71}

\newcommand{\coordN}{129}

\newcommand{\coordTranslatedCloses}{126}

\newcommand{\XoneNullTypes}{4}
\newcommand{\XoneNullEffTypes}{3.79}
\newcommand{\XthreeNullTypes}{3}
\newcommand{\XthreeNullEffTypes}{2.59}
\newcommand{\XtwoNullTypes}{3}
\newcommand{\XtwoNullEffTypes}{2.83}

\newcommand{\ShapeProblems}{24}
\newcommand{\ShapeTraj}{120}
\newcommand{\ShapeNew}{19}
\newcommand{\ShapeReturn}{0}
\newcommand{\ShapePosAUROC}{0.39}
\newcommand{\ShapeEffLo}{1.0}
\newcommand{\ShapeEffHi}{1.3}
\newcommand{\ShapeNullLo}{0.52}
\newcommand{\ShapeNullHi}{0.59}

\newcommand{\XthreeConfigs}{3}
\newcommand{\XthreeTriples}{288}

\newcommand{\XthreeNullEvals}{15}

\newcommand{\XthreeNullGone}{12}
\newcommand{\XthreeNARstructural}{0.80}

\newcommand{\XthreeRtwoPerConfig}{0, 0, 0}
\newcommand{\XthreeCkaPerConfig}{4, 4, 4}

\newcommand{\XthreePAR}{1.00}

\newcommand{\XthreeProteins}{320}
\newcommand{\XthreeResidues}{128}
\newcommand{\XoneConfigs}{4}
\newcommand{\XoneTriples}{480}

\newcommand{\XoneNullEvals}{20}

\newcommand{\XoneNullGone}{12}
\newcommand{\XoneNARstructural}{0.60}

\newcommand{\XoneNARinsample}{0.80}

\newcommand{\XoneSweepMax}{8{,}192}
\newcommand{\XoneNARinsampleLarge}{0.20}

\newcommand{\XoneRtwoPerConfig}{0, 0, 1, 1}
\newcommand{\XoneCkaPerConfig}{4, 0, 4, 4}
\newcommand{\XoneNullsPerConfig}{5}
\newcommand{\XoneWorstNullRaw}{0.23}
\newcommand{\XoneWeakestRealRaw}{0.67}

\newcommand{\XoneRealAccepted}{4}

\newcommand{\XoneSeeds}{5}
\newcommand{\XonePos}{8}
\newcommand{\XoneDonors}{12}
\newcommand{\XoneRecip}{16}
\newcommand{\XoneCausalMin}{0.2}

\newcommand{\XtwoFeatures}{405}
\newcommand{\XtwoCAR}{0.45}

\newcommand{\XtwoUnsupportedPct}{55\%}
\newcommand{\XtwoControl}{0.005}
\newcommand{\XtwoDelta}{+0.45}

\newcommand{\XtwoNAR}{0.00}
\newcommand{\XtwoDeltaCIlo}{+0.29}
\newcommand{\XtwoDeltaCIhi}{+0.58}

\newcommand{\XtwoNfeat}{150}
\newcommand{\XtwoCurveK}{10, 25, 50, 100, 150}
\newcommand{\XtwoCurve}{0.55, 0.55, 0.52, 0.45, 0.45}
\newcommand{\XtwoFlowLo}{605}
\newcommand{\XtwoFlowHi}{1453}
\newcommand{\XtwoUntestable}{191}
\newcommand{\XtwoNoControl}{4}

\newcommand{\XtwoEnrTestable}{55.4}
\newcommand{\XtwoEnrUntestable}{52.6}

\newcommand{\XtwoNullN}{16}

\newcommand{\XtwoAblNegN}{222}
\newcommand{\XtwoActOnAblNeg}{100}
\newcommand{\XtwoEither}{0.70}
\newcommand{\XtwoNeither}{0.30}

\newcommand{\XtwoCARvalid}{0.40}
\newcommand{\XtwoControlValid}{0.000}

\newcommand{\XtwoSelected}{600}
\newcommand{\XtwoCARactivate}{0.65}

\newcommand{\XtwoAgreeN}{162}
\newcommand{\XtwoAblationN}{183}

\newcommand{\XtwoPositiveOnValid}{0.90}

\newcommand{\XtwoControlFirst}{0.02}
\newcommand{\XtwoDeltaFirst}{+0.43}
\newcommand{\XtwoDeltaFirstCIlo}{+0.27}
\newcommand{\XtwoDeltaFirstCIhi}{+0.57}

\newcommand{\XtwoSurrLo}{1.0}
\newcommand{\XtwoSurrHi}{1.2}
\newcommand{\XtwoTopEnrLo}{137}
\newcommand{\XtwoTopEnrHi}{187}
\newcommand{\XtwoNullEnrHi}{1.2}
\newcommand{\XtwoCARallSelected}{0.31}
\newcommand{\XtwoSpecN}{600}
\newcommand{\XtwoSpecCAR}{0.46}
\newcommand{\XtwoSpecControl}{0.07}
\newcommand{\XtwoSpecDelta}{+0.39}
\newcommand{\XtwoSpecArmsPositive}{4}
\newcommand{\XtwoSpecArms}{4}
\newcommand{\XtwoSpecNAR}{0.06}

\newcommand{\XtwoSpecNullPassEnr}{69}

\newcommand{\LfourOffsetA}{2.5}
\newcommand{\LfourOffsetB}{5}

\newcommand{\LfourHiddenZa}{16}
\newcommand{\LfourHiddenZb}{31}
\newcommand{\LfourHiddenA}{2.5}
\newcommand{\LfourHiddenB}{5}
\newcommand{\LfourThresholdPct}{6}

\newcommand{\posThetaBySep}{$0.05$ rad at $62.9\times$, $0.08$ rad at $25.2\times$ and $12.6\times$, and $0.03$ rad at $6.3\times$}
\newcommand{\XtwoUntestableZero}{141}
\newcommand{\XtwoUntestableSome}{50}
\newcommand{\XtwoUntestableMaxPos}{24}
\newcommand{\XtwoTestableMinPos}{31}
\newcommand{\XtwoMinOn}{30}
\newcommand{\XtwoTopConceptN}{118}
\newcommand{\XtwoTopConceptGuard}{140}
\newcommand{\XtwoTopConceptGuardSeq}{47}
\newcommand{\XtwoTopConceptCons}{0}

\begin{document}

\maketitle

\begin{abstract}
Automated discovery systems propose scientific explanations and test which to accept. Does passing a test establish the intended claim? We develop four executable operations for changing physical descriptions: identifying redundancies (L1), coarse-graining dynamics (L2), transferring structure (L3), and inferring missing quantities from conservation imbalances (L4). We evaluate these operations and three interpretability settings through null acceptance (NAR), positive acceptance (PAR), and consequence agreement (CAR).

A reduction mapping every state to zero predicts its future perfectly. It passes closure on \dystsConstPct\ of \dystsN\ public chaotic systems despite retaining no dynamical variation. An observable-preservation test rejects it throughout; together the criteria accept 24/24 genuine reductions and reject 72/72 nulls. Other operations distinguish symmetries from redundancies, validate transfers, and reject spurious conservation imbalances. In language and ESM-2 protein models, linear CKA remains unchanged when a proposed correspondence is randomly rotated, whereas localized interventions distinguish genuine mappings from tested nulls. In sparse autoencoders, enrichment rejects all \XtwoNullN\ tested nulls, yet ablation supports \XtwoCAR\ of evaluable enriched features compared with \XtwoControl\ of matched controls.

These results reveal two acceptance gaps: descriptive tests can accept scientifically empty candidates, and accepted candidates can lack independently verified consequences. Reliable automated discovery requires evaluating both what a test accepts and the scientific consequences it establishes.

\ifanon
Code and data are provided in the supplementary material.
\else
Code and data:

\url{https://real-project-aether.github.io/project-aether/}.
\fi
\end{abstract}


\section{Introduction}

Automated scientific discovery increasingly relies on AI systems to propose hypotheses, construct models, and evaluate their results. Yet proposing a candidate is only part of discovery. A system must also decide whether the candidate supports a meaningful scientific claim. This decision often relies on a test suggested by the candidate's defining property: a reduced physical model is accepted when its dynamics close, two neural representations are considered similar when their alignment score is high, and a learned feature is assigned a meaning when its activations correlate with that concept. These tests establish measurable properties, but the scientific conclusions drawn from them may require additional evidence.

Consider the problem of discovering a reduced description of a dynamical system. A standard requirement is \emph{closure}: the reduced variables must predict their own future without access to the original state \citep{zwanzig1960ensemble}. Now consider the constant reduction $Q(x)=0$, which maps every state to zero. Its dynamics are exactly $\dot q=0$, so it predicts its own future perfectly, regardless of the original system. The reduction achieves zero closure error while preserving no dynamical variation. \textbf{A scientifically empty description can therefore satisfy the defining criterion of a successful reduction.} This exposes a fundamental question: \emph{what does passing an acceptance test actually establish about a scientific discovery?}

We investigate this question by developing four executable operations for changing scientific descriptions. Motivated by recurring forms of conceptual change in historical discoveries (Appendix~\ref{sec:corpus}), these operations identify and reinterpret unobservable degrees of freedom (\textbf{L1}), construct reduced dynamical descriptions (\textbf{L2}), transfer structure between physical systems (\textbf{L3}), and infer missing quantities from conservation-law imbalances (\textbf{L4}). Each operation transforms an existing description into a candidate scientific revision and evaluates the conditions under which that revision is justified. Together, they provide computational mechanisms for proposing distinct forms of conceptual change, making it possible to examine how scientific claims are generated and accepted.

The central challenge is determining whether a candidate's defining property establishes its intended scientific consequence. We distinguish a \emph{descriptive test}, which measures properties such as closure or structural agreement, from a \emph{consequence test}, which evaluates an independently specified implication of the proposed revision. Across the four operations, these consequences include preserving physical observables, distinguishing redundancies from conserved symmetries, validating transferred conservation laws, and establishing that a measured imbalance exceeds the relevant noise scale. We evaluate both tests using controlled candidates whose intended scientific content is known.

The results reveal why this distinction matters. On the public \texttt{dysts} benchmark \citep{gilpin2021chaos}, the constant reduction passes closure on \dystsConstCloses\ of \dystsN\ chaotic systems, while an observable-preservation test rejects it throughout. In constructed slow--fast systems, the combined criteria accept all 24 known reductions and reject all 72 null candidates. The remaining operations distinguish four conserved symmetries from a gauge redundancy, validate six genuine transfers while rejecting six broken-symmetry controls, and recover a missing charge of $1.192$ against a withheld value of $1.200$. These experiments demonstrate the operations' ability to produce and evaluate candidate scientific revisions, while exposing the limitations of their descriptive acceptance criteria.

We next ask whether the same problem extends to neural interpretability. Linear centered kernel alignment (CKA), widely used to compare neural representations \citep{kornblith2019similarity,raghu2021vit,davari2023cka}, is independent of any proposed correspondence between them. A randomly rotated mapping can therefore retain the original CKA score while breaking the claimed correspondence. We demonstrate this ambiguity in Pythia language models and ESM-2 protein language models, where localized interventions distinguish genuine mappings from tested nulls by measuring whether the mappings transfer functional effects.

A second problem emerges in sparse-autoencoder interpretability. Learned features are often interpreted through concepts enriched among their activations \citep{bricken2023monosemanticity}. We find that enrichment rejects all tested null features, yet only a subset of accepted features receives concept-specific support from independent ablation experiments. Matched controls show that enrichment carries useful information, while positive controls reveal the sensitivity of the intervention test itself. Thus, rejecting empty candidates does not by itself establish the scientific interpretation of those accepted.

\begin{figure}[t]
\centering
\begin{tikzpicture}[
  font=\small,
  node distance=5mm and 8mm,
  box/.style={
    draw,
    rounded corners=2pt,
    align=center,
    inner sep=4pt,
    minimum height=13mm
  },
  audit/.style={
    draw,
    dashed,
    rounded corners=2pt,
    align=center,
    inner sep=4pt,
    font=\footnotesize
  },
  ar/.style={
    -{Stealth[length=2mm]},
    semithick
  }
]

\node[box, text width=23mm] (cand)
  {\textbf{Candidate}\\
   {\footnotesize scientific revision}};

\node[box, right=of cand, text width=29mm] (guard)
  {\textbf{Descriptive test}\\
   {\footnotesize Does the defining property hold?}};

\node[box, right=8mm of guard, text width=32mm] (cons)
  {\textbf{Consequence test}\\
   {\footnotesize Does the scientific implication hold?}};

\node[box, right=8mm of cons, text width=19mm] (acc)
  {\textbf{Supported}\\
   {\footnotesize scientific claim}};

\draw[ar] (cand) -- (guard);
\draw[ar] (guard) -- node[
  above,
  font=\scriptsize,
  fill=white,
  inner sep=1pt
] {$G(c)=1$} (cons);
\draw[ar] (cons) -- node[
  above,
  font=\scriptsize,
  fill=white,
  inner sep=1pt
] {$H(c)=1$} (acc);

\node[audit, below=9mm of guard, text width=32mm] (a1)
  {\textbf{Audit descriptive acceptance}\\[2pt]
   Null candidates $\rightarrow$ NAR\\
   Known positives $\rightarrow$ PAR};

\node[audit, below=9mm of cons, text width=36mm] (a2)
  {\textbf{Audit scientific consequences}\\[2pt]
   Accepted candidates $\rightarrow$ CAR\\
   Positive and matched controls};

\draw[ar, dashed] (a1.north) -- (guard.south);
\draw[ar, dashed] (a2.north) -- (cons.south);

\end{tikzpicture}

\caption{
\textbf{Generating and evaluating scientific revisions.}
The four physical discovery operations (L1--L4) construct candidate
changes to scientific descriptions. A descriptive test $G$ evaluates
each candidate's defining property, while an independent consequence
test $H$ evaluates the scientific implication of accepting it.
We measure null acceptance (NAR), positive acceptance (PAR),
and consequence agreement (CAR), with positive and matched controls
used to assess consequence-test reliability. The same evaluation
principle is applied to neural representation correspondence and
sparse-autoencoder interpretation.
}
\label{fig:audit}
\end{figure}
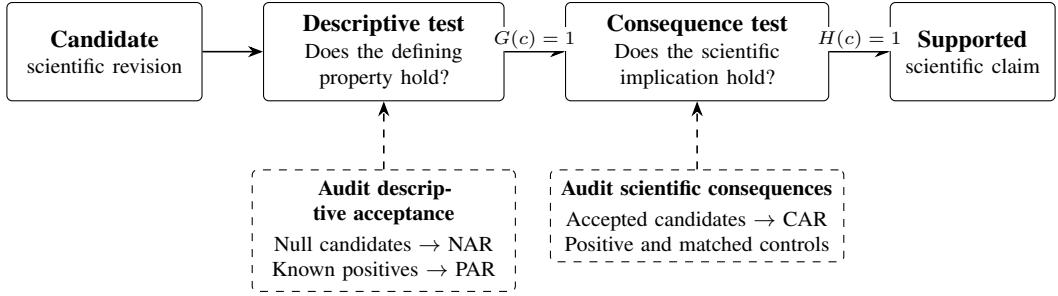

Together, these experiments identify two gaps between descriptive acceptance and scientific evidence: tests can accept candidates that carry none of the intended information, and accepted candidates can lack independent support for their proposed consequences. We quantify these gaps through the \emph{null acceptance rate} (NAR), \emph{positive acceptance rate} (PAR), and \emph{consequence agreement rate} (CAR) (Fig.~\ref{fig:audit}). Our contribution connects the construction of scientific descriptions with the evidence required to accept them: \textbf{scientific discovery becomes reliable when candidate revisions are evaluated through the consequences that give their claims meaning.}

\section{Results}
\label{sec:results}

\subsection{An empty model can satisfy dynamical closure}
\label{sec:guard}

Discovering a reduced description of a dynamical system requires identifying which state variables are sufficient to predict its future. A standard requirement is \emph{closure}: the reduced variables must evolve according to a law that depends only on those variables, without requiring the original state \citep{zwanzig1960ensemble}. Successful closure is commonly interpreted as evidence that the reduced description captures the relevant dynamics.

Consider the simplest possible reduction,
\begin{equation}
    Q(x)=0.
\end{equation}
Every state maps to the same constant, so the reduced trajectory is $q(t)=0$ and its dynamics are exactly $\dot q=0$. The coarse law predicts this trajectory with zero error, regardless of how complicated or chaotic the original dynamics may be. The reduction therefore satisfies closure perfectly while preserving no variation in the original system. \textbf{The very act of discarding all information makes the reduced dynamics perfectly predictable.}

This is a structural ambiguity in the acceptance criterion. Closure measures whether the chosen representation predicts its own evolution; it does not measure how much of the original system that representation describes. Additional observations cannot resolve this ambiguity because the constant trajectory remains perfectly predictable for every initial condition and prediction horizon.

We asked whether this mathematical counterexample also passes a practical model-evaluation pipeline. We evaluated it on the public \texttt{dysts} benchmark \citep{gilpin2021chaos}, which contains 137 published chaotic dynamical systems. Of these, \dystsN\ supported our finite-dimensional evaluation. The same closure criterion and threshold ($0.25$) used in our controlled experiments were applied without retuning. The coarse law was fitted on trajectories separate from those used for evaluation.

\begin{figure}[t]
\centering
\includegraphics[width=0.88\linewidth]{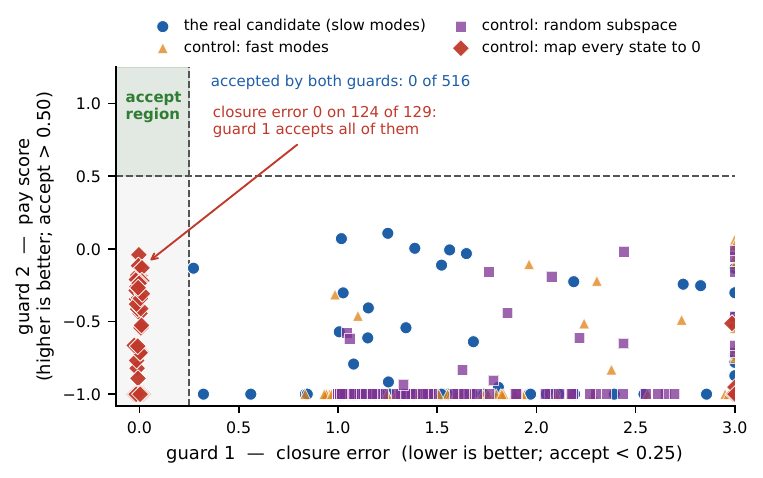}
\caption{
\textbf{An empty reduction passes dynamical closure.}
On \dystsN\ external chaotic systems, the constant reduction satisfies closure in \dystsConstCloses\ cases despite discarding all dynamical variation. An independent observable-preservation criterion rejects it in every case. Both thresholds were fixed before evaluation. Non-finite closure errors are counted as failures.
}
\label{fig:external}
\end{figure}

The constant reduction passed on \textbf{\dystsConstCloses\ of \dystsN\ systems}, achieving exactly zero closure error (Fig.~\ref{fig:external}). The remaining \dystsConstNonfinite\ cases were recorded as failures because lifting the constant state to the fine-state origin encountered an undefined right-hand side. This numerical issue does not remove the underlying ambiguity: a zero-dimensional formulation requiring no lifted state closes on all \dystsN\ systems (Appendix~\ref{app:external}).

Thus, a criterion designed to identify autonomous reduced dynamics accepts a representation that contains none of the system's dynamical variation. The failure persists on independently published systems and under held-out evaluation. It arises from what closure measures, rather than from overfitting or insufficient optimization.

The result raises a more precise scientific question: \emph{what must a reduced model preserve for its predictive success to describe the original system?} We address this by testing the information retained about independently specified observables.

\subsection{Independent consequences distinguish meaningful reductions}

Closure evaluates whether reduced variables predict themselves. To determine whether they retain information about the original system, we introduce an independent requirement: preservation of specified observables.

Let $P(x)$ denote observables chosen before evaluation, and $Q^\dagger Qx$ the state reconstructed from reduced coordinates. We measure preservation using
\begin{equation}
\label{eq:pay}
\mathrm{Pay}(Q)=
\operatorname{clip}_{[-1,1]}
\left(
1-
\frac{
\sum_{x\in\mathcal E}
\|P(Q^\dagger Qx)-P(x)\|_{\Sigma}^{2}
}{
\sum_{x\in\mathcal E}
\|P(x)-\bar P\|_{\Sigma}^{2}
}
\right),
\end{equation}
where $\mathcal E$ is an ensemble of states, $\bar P$ and $\Sigma$ are the mean and covariance of the observables, and
$\|v\|_\Sigma^2=v^\top\Sigma^{-1}v$.
The score equals one for an identity reduction and is nonpositive for a constant reconstruction. We require $\mathrm{Pay}(Q)>0.5$ alongside closure.

On a controlled slow--fast system, constant and fast-mode reductions both achieve perfect closure but score $-0.54$ and $-0.50$ on observable preservation. On the external \texttt{dysts} benchmark, the preservation test rejects the constant reduction on all \dystsN\ systems.

We then tested whether the combined criteria retain genuine reductions. Across 24 constructed slow--fast systems with known reduced subspaces, closure alone accepts 47 of 72 null candidates. The combined criteria accept \textbf{0 of 72 nulls and 24 of 24 true subspaces}. Using subspaces estimated by the proposer reduces positive acceptance to 17 of 24. These controlled tests use derived coarse dynamics, while the external benchmark uses learned dynamics. The linear proposer finds no accepted nontrivial reduction on the external chaotic systems, limiting the demonstrated recovery capability (Appendix~\ref{app:external}).

Together, these experiments show that independently specified observables distinguish informative reductions from the tested empty solutions.

\paragraph{Measuring scientific acceptance.}
\label{sec:framework}
We formalize this distinction through a descriptive test $G(c)$ and an independent consequence test $H(c)$ for candidate $c$. We measure three quantities:
\begin{align}
\widehat{\mathrm{NAR}}
&=\frac{1}{|\mathcal V|}
\sum_{c\in\mathcal V}G(c),
\label{eq:nar}\\
\widehat{\mathrm{PAR}}
&=\frac{1}{|\mathcal P|}
\sum_{c\in\mathcal P}G(c),
\label{eq:par}\\
\mathrm{CAR}
&=\Pr_{c\sim\mathcal C}
[H(c)=1\mid G(c)=1,\ c\notin\mathcal V].
\label{eq:car}
\end{align}
Here $\mathcal V$ and $\mathcal P$ are specified null and positive candidate suites, and $\mathcal C$ is the evaluated candidate distribution. The \emph{null acceptance rate} (NAR) measures acceptance of candidates lacking the intended information; the \emph{positive acceptance rate} (PAR) checks whether genuine candidates are recognized; and the \emph{consequence agreement rate} (CAR) measures independent support among accepted candidates. Because CAR depends on the sensitivity of $H$, we also test $H$ using positive and matched controls.

The coarse-graining experiments reveal how an accepted physical description can lose its intended content. We next examine whether the same problem arises across other ways of revising a scientific theory.

\subsection{Four executable operations for revising scientific descriptions}
\label{sec:ops}

We implement four operations motivated by recurring forms of conceptual change in science (Appendix~\ref{sec:corpus}). Each takes an existing physical description and constructs a candidate revision: removing redundant parameters (L1), changing dynamical variables (L2), transferring structure between systems (L3), or introducing a missing conserved quantity (L4). Each revision has a descriptive criterion and a test of its intended physical consequence (Table~\ref{tab:guard}).

\textbf{L1: Reinterpretation.} We identify parameter directions that leave measured predictions invariant using the null space of their Jacobian. On a constructed lattice gauge model, five directions appear unobservable. Testing transformations locally distinguishes one gauge redundancy from four global symmetries, preserving four conserved charges (maximum drift $7.6\times10^{-10}$). Uniform invariance alone would incorrectly discard all five directions.

\textbf{L2: Coarse-graining.} We construct reduced state variables and test their dynamical closure. As established above, an empty reduction can close perfectly. Requiring preservation of independent observables rejects the tested empty reductions while retaining known true subspaces.

\textbf{L3: Transfer.} We fit a map between two physical systems and propose that a source-system conservation law also holds in the target. Structural agreement alone accepts two of six deliberately broken-symmetry controls on held-out states. Testing the transferred charge under the target's own dynamics accepts all six genuine transfers and rejects all six controls.

\textbf{L4: Missing quantities.} We infer a missing conserved quantity from the imbalance between measured inputs and outputs. A significance-only criterion responds to small calibration offsets even when nothing is missing. A noise-scale requirement rejects those cases, while a positive control recovers a hidden charge of $1.192$ per event against the withheld value $1.200$.

\begin{table}[t]
\centering
\caption{\textbf{Four executable operations and the scientific consequences used to evaluate them.} Descriptive criteria admit misleading candidates; independent tests recover the intended distinctions in controlled cases. Full methods and sensitivity analyses appear in Appendices~\ref{app:ops} and~\ref{app:results}.}
\label{tab:guard}
\footnotesize
\setlength{\tabcolsep}{3pt}
\begin{tabular}{@{}l p{0.23\linewidth} p{0.27\linewidth} p{0.34\linewidth}@{}}
\toprule
 & Candidate revision & Descriptive ambiguity & Independent consequence \\
\midrule
L1 & Remove unobservable directions & Global symmetry resembles redundancy & Locality separates 1 redundancy from 4 conserved symmetries \\
L2 & Reduce dynamical variables & Empty states satisfy closure & Preserve observables; 0/72 nulls and 24/24 true subspaces accepted \\
L3 & Transfer structure across systems & 2/6 broken controls match structurally & Target dynamics validate 6/6 transfers, reject 6/6 controls \\
L4 & Infer an unobserved quantity & Calibration offsets appear significant & Noise-scale test rejects offsets; recovers $1.192$ vs. $1.200$ \\
\bottomrule
\end{tabular}
\end{table}

These operations make different kinds of scientific revision executable. Their controlled results also reveal a shared requirement: a successful descriptive test must be connected to the physical consequence that gives the revision its scientific meaning. We next test this principle on representations learned by neural networks.

\subsection{Representation similarity does not certify correspondence}
\label{sec:xai}

Representational similarity is widely used to compare neural networks \citep{kornblith2019similarity,raghu2021vit,davari2023cka}. However, similarity between two representation spaces does not establish that a particular mapping between them is correct.

We fit a linear map $W$ between model representations and evaluate it using held-out reconstruction $R^2$ and linear CKA, accepting when either exceeds $0.5$. CKA depends on the two representation sets, independently of the proposed map. Consequently, composing $W$ with a random rotation can destroy the correspondence while leaving CKA unchanged. In our experiments, such a rotation reduces $R^2$ to approximately $-0.96$ without changing CKA.

To test the claimed correspondence directly, we perturb the residual representation of model A at one position, transfer the perturbation through $W$ to model B, and compare their induced output changes. A genuine correspondence should transfer these effects consistently.

Across $\XoneConfigs$ Pythia model-layer configurations \citep{biderman2023pythia}, the descriptive criterion accepts \XoneNullGone\ of \XoneNullEvals\ null evaluations on held-out data. The intervention test rejects every tested null while accepting the genuine correspondence in all \XoneRealAccepted\ configurations.

We repeat the same evaluation on ESM-2 protein language models \citep{lin2023esm2}, where interventions are assessed through masked-residue predictions. The descriptive criterion accepts \XthreeNullGone\ of \XthreeNullEvals\ null evaluations, whereas the combined criteria reject all tested nulls. Identity and adjacent-layer positive controls are accepted in all six evaluations ($\mathrm{PAR}=\XthreePAR$). Additional sampled null families show the same pattern (Appendix~\ref{app:xai}).

Thus, representational similarity can leave a proposed mapping unverified even on held-out data. Testing whether the mapping transfers functional effects resolves this ambiguity for the correspondences examined here.

This leads to a second question: if a descriptive test rejects every null, does that establish the scientific interpretation of the candidates it accepts?

\subsection{Feature enrichment does not guarantee causal specificity}

The preceding experiments show that descriptive tests can accept incorrect or empty candidates. An apparent solution is to challenge every acceptance criterion with null examples and retain only criteria that reject them. But does successful null rejection establish the scientific meaning of the candidates that pass?

We investigate this question using sparse autoencoders (SAEs), which decompose neural activations into learned features. These features are commonly interpreted by examining their strongest activating examples \citep{bricken2023monosemanticity}. If a feature activates disproportionately on examples associated with a concept, that enrichment provides evidence of an association. A stronger interpretation is that the feature contributes specifically to the model's behavior on that concept. We ask how often the first observation supports the second.

We train TopK autoencoders on four Pythia model-layer configurations: layers 4 and 8 of Pythia-160M, and layers 8 and 16 of Pythia-410M. Each autoencoder uses an $8\times$ expansion and $L_0=32$, recovering 93--99\% of activation variance. The data are divided into three disjoint splits (60/20/20) for autoencoder training, feature interpretation, and intervention testing. This ensures that the examples used to assign a concept to a feature are separate from those used to test its effect.

\paragraph{Enrichment rejects simple null features.}
We associate each feature with the next-token concept most enriched among its activations, relative to that token's base frequency. The descriptive test accepts features with enrichment greater than four. We challenge it with four null constructions per model-layer configuration: a random direction, a coordinate-scrambled feature, a noisy surrogate of a learned feature, and a near-dead feature.

All \XtwoNullN\ null evaluations fail the enrichment threshold ($\mathrm{NAR}=\XtwoNAR$). Thus, the descriptive test distinguishes strongly enriched features from these controls. The null suite is limited, however: even the noisy surrogate retains relatively little of the source feature's enrichment. Rejection of these candidates therefore establishes a useful initial check, while leaving the interpretation of accepted features unresolved.

\paragraph{Strong enrichment provides incomplete causal support.}
We select the $\XtwoNfeat$ most enriched features from each configuration and evaluate them on the independent consequence split. Because selection is based on enrichment, descriptive acceptance is guaranteed for these candidates. The meaningful question is whether their associated concepts are supported by a separate functional measurement.

For a feature $f$ associated with concept $C$, we subtract its contribution from the model's residual representation and measure the change in prediction loss. A concept-specific feature should affect prediction of $C$ more strongly than prediction elsewhere. We define
\begin{equation}
S_f=
\frac{
\Delta L_C-\Delta L_{\neg C}
}{
\mathrm{SE}[\Delta L_C-\Delta L_{\neg C}]
},
\end{equation}
where $\Delta L_C$ and $\Delta L_{\neg C}$ denote loss increases following ablation on positions associated with $C$ and on other positions, respectively. The standard error is calculated from per-sequence contrasts. We count a feature as supported when $S_f>1.96$ and $\Delta L_C>0$.

Ablating an active feature may change predictions even when its effect is unrelated to the proposed concept. To distinguish concept-specific effects from general perturbation effects, we construct three matched controls for each selected feature. Controls are drawn from the same model and layer, have similar firing rates and activation magnitudes, and are associated with different top concepts. They are evaluated on the selected feature's concept under the same intervention criterion.

Of the \XtwoSelected\ selected features, \XtwoUntestable\ lack sufficient concept occurrences in the consequence split, and \XtwoNoControl\ lack an eligible matched control. The paired evaluation therefore contains \XtwoFeatures\ features.

The intervention supports $\XtwoCAR$ of the selected features, compared with $\XtwoControl$ of matched controls (Table~\ref{tab:x2car}). The difference,
$\Delta\mathrm{CAR}=\XtwoDelta$,
has a 95\% confidence interval of $\XtwoDeltaCIlo$ to $\XtwoDeltaCIhi$, and is positive across all four configurations. The interval accounts for clustering among features sharing the same concept.

This result establishes two properties simultaneously. Strong enrichment carries information about concept-specific effects beyond what is explained by matched feature activity. Yet the descriptive criterion does not guarantee that an accepted feature satisfies the corresponding intervention test. \textbf{A feature can have a meaningful statistical association with a concept without establishing the particular causal role attributed to it.}

\begin{table}[t]
\centering
\caption{
\textbf{Enrichment predicts causal specificity imperfectly.}
Consequence agreement for strongly enriched SAE features and activity-matched controls, evaluated on the same concepts. The overall difference is $\Delta\mathrm{CAR}=\XtwoDelta$ (95\% CI $\XtwoDeltaCIlo$--$\XtwoDeltaCIhi$).
}
\label{tab:x2car}
\small
\begin{tabular}{@{}lrrrr@{}}
\toprule
Arm & Paired & Selected CAR & Control CAR & $\Delta$CAR \\
\midrule
Pythia-160m L4 & 93 & 0.24 & 0.00 & $+0.24$ \\
Pythia-160m L8 & 111 & 0.54 & 0.00 & $+0.54$ \\
Pythia-410m L16 & 102 & 0.58 & 0.00 & $+0.58$ \\
Pythia-410m L8 & 99 & 0.42 & 0.02 & $+0.40$ \\
\midrule
pooled & \textbf{405} & \textbf{0.45} \, {\footnotesize (CI $0.29$--$0.58$)} & \textbf{0.005} & $\mathbf{+0.45}$
 \\
\bottomrule
\end{tabular}
\end{table}

\paragraph{The consequence test must itself be validated.}
Incomplete intervention support admits two explanations: some enriched features lack the proposed functional specificity, or the intervention test fails to detect effects that are present. To assess its sensitivity, we construct positive controls using the gradient of the concept token's logit with respect to the residual representation. These directions provide interventions expected to influence prediction of the selected concept.

The controls reveal that the test's sensitivity varies across configurations. In the two Pythia-410M arms, gradient-direction ablation frequently produces the wrong relative effect, preventing those controls from validating the test. In the two Pythia-160M arms, the controls satisfy the specified validity criterion, and the intervention test detects $\XtwoPositiveOnValid$ of the defined directions. Within these configurations, consequence agreement for enriched features is $\XtwoCARvalid$, compared with $\XtwoControlValid$ for matched controls.

We additionally test the complementary effect of activating a feature where it is normally silent. This supports $\XtwoCARactivate$ of paired features, with either intervention supporting $\XtwoEither$ (Appendix~\ref{app:xai}). These experiments show that functional support depends partly on the consequence being tested. Accordingly, failure to satisfy one intervention criterion does not establish that a feature is causally irrelevant.

The SAE experiment exposes a second acceptance gap. A descriptive test may reject every candidate in its null suite while leaving the functional interpretation of accepted candidates only partially supported. Evaluating scientific acceptance therefore requires both testing what a criterion rejects and measuring the independent consequences of what it accepts.

\section{Related Work}
\label{sec:related}

Related concerns have appeared in interpretability research. Sanity checks reveal explanations that survive weight randomization \citep{adebayo2018sanity}; probing selectivity evaluates performance against control tasks \citep{hewitt2019control}; and causal interventions test the effects of internal representations \citep{geiger2021causal,meng2022rome}. Prior work also identifies limitations of representational similarity \citep{davari2023cka}, while SAEBench and MIB provide standardized interpretability evaluations \citep{karvonen2025saebench,mueller2025mib}. In automated discovery, verifying recovered structure remains an open challenge \citep{posner2026mwm}, and candidate-selection systems increasingly determine which scientific claims advance \citep{foster2026rpm,zheng2026far}. We connect these concerns through a common empirical question: what scientific consequences are established by a candidate's acceptance?

\section{Discussion}
\label{sec:discussion}

We contribute four executable operations for revising physical descriptions and a common way to evaluate the scientific claims they produce. The operations identify redundancies, change dynamical variables, transfer conserved structure, and infer missing quantities. Their controlled evaluations, together with the neural interpretability experiments, show that descriptive acceptance and scientific support can diverge: closure can certify an empty reduction, similarity can accept an incorrect mapping, and enrichment can leave concept-specific effects unresolved.

The experiments identify two complementary requirements. \emph{Null controls} test whether an acceptance criterion distinguishes meaningful candidates from empty or misleading ones. \emph{Independent consequence tests} evaluate the scientific implications of accepted candidates, with positive controls establishing what those tests can detect. NAR, PAR, and CAR quantify these properties within explicitly specified candidate distributions.

The general principle is simple: \textbf{an accepted scientific candidate should support a consequence that gives its interpretation empirical content.} Connecting executable changes of scientific description to independently tested consequences provides a practical foundation for more reliable automated discovery.

\section{Limitations}
\label{sec:limits}

Our null suites are finite and selected by us; low NAR does not guarantee rejection of other nulls. Consequence tests have limited sensitivity, as illustrated by the SAE positive controls. The physics proposers use linear maps and cannot recover the curved reductions in the external chaotic benchmark. Experiments use Pythia models up to 410M parameters and ESM-2 up to 150M. The L1, L3, and L4 demonstrations use constructed physical systems; their generality to open-ended scientific problems remains untested. The historical prize corpus motivated the operations but has only one annotator (Appendix~\ref{sec:corpus}).


\bibliographystyle{plainnat}
\bibliography{refs}

\appendix

\section{The four operations in full}
\label{app:ops}

\paragraph{L1, reinterpretation.} For a theory with parameters $p$ predicting $f(x,p)$, the
directions along which no prediction moves are the null space of $J_{ij}=\partial f_i/\partial
p_j$: structural non-identifiability~\citep{raue2009identifiability}, read as a property of a
theory rather than a nuisance in a fit. The obvious move is to quotient that null space away.
\emph{This is wrong}: a global phase rotation is invisible to every probe exactly as a gauge
transformation is, and removing both erases the conserved quantity instead of discovering it.
Three tests must be kept apart --- \textbf{probe invariance}, \textbf{dynamical invariance}, and
the \textbf{local versus uniform} distinction that separates gauge redundancy from a global
symmetry (Table~\ref{tab:locality}). We do not claim probe invariance yields a Noether charge in
general; within this constructed model the local test is what classifies one.

\paragraph{L1 on data with error bars.} Exact degeneracy is the wrong test for real data: Lorentz's
aether was not unobservable in principle, only to within the precision of the day. We scale column
$j$ of the Jacobian by $|p_j|$ and row $i$ by $1/\sigma_i$, and ask whether a direction moves by an
order-unity amount without shifting any prediction past its error bar. The second guard is the
locality test of Table~\ref{tab:locality}.

\begin{table}[h]
\caption{The locality test: one null-space computation, two different moves. This is the second
guard for L1, and the difference between deleting a structure and discovering a conservation law.}
\label{tab:locality}
\centering\small
\begin{tabular}{@{}ll@{}}
\toprule
invariant \emph{point by point} & redundancy: quotient it \quad [L1] \\
invariant only \emph{uniformly} & symmetry: keep it, read off a conserved charge \quad [L4] \\
\bottomrule
\end{tabular}
\end{table}

A direction that no probe sees when it is moved uniformly, but that probes do see when it is moved
locally, is a global symmetry and carries a conserved charge. The implementation refuses to quotient at all when no locality test is supplied:
\texttt{quotient()} raises rather than proceeding, because the failure mode it guards against is
silent.

\paragraph{L2, coarse-graining.} Closure is scored by integrating the coarse law alone and
comparing with the truth, normalised by the trajectory's own spread. The \emph{derived} law
projects the known fine law through the reduction. The \emph{learned} law is a polynomial regression
from coarse states to coarse velocities, with velocity targets taken from the fine law at the lifted
state. It has linear and cubic terms on the lattice and terms up to quadratic on the benchmark. Which law carries the guard depends
on the experiment. On the slow--fast systems (Table~\ref{tab:vacuity}, the separation sweep and
the labelled suite of App.~\ref{app:external}) closure uses the derived law, which has nothing
fitted. On the lattice both laws are computed and the better one is used. The learned law there is
fitted on every trajectory, including the scored one. Both choices favour acceptance, and the
lattice reduction is refused under both. On the external benchmark acceptance rests on the learned
law alone, fitted on trajectories other than the scored one. With both laws reported, a negative
can be traced to the reduction or to the fitter. The pay guard is fit-free by
construction, after three fitted versions failed in three different ways, none of which produced an
obviously wrong number. Its hard invariant is that a reduction keeping everything must score
exactly $1.000$.

\begin{table}[h]
\caption{Three vacuity controls on the slow--fast system at $\epsilon=0.02$. Two of the three close
\emph{perfectly}; only the pay guard of Eq.~\ref{eq:pay} refuses them, far below its $0.50$
threshold. Pay values are from the covariance-weighted score; the unweighted version this replaced
drove both empty cases to the $-1$ clip, which reported saturation rather than a measurement.}
\label{tab:vacuity}
\centering\small
\begin{tabular}{@{}lccl@{}}
\toprule
control & closes & pays & refused by \\
\midrule
constant reduction (every state to zero) & \textbf{0.000} & $-0.54$ & guard 2 only \\
random subspace & 2.167 & $0.18$ & guard 1 \\
the \textbf{fast} modes instead of the slow ones & \textbf{0.000} & $-0.50$ & guard 2 only \\
\bottomrule
\end{tabular}
\end{table}

\paragraph{Properties of the pay score.} Write $R=Q^{\dagger}Q$ for the reconstruction and
$\mathcal{E}$ for the state ensemble. Three properties follow directly from Eq.~\ref{eq:pay}, and
the first and third are asserted in the shipped code.

\begin{enumerate}\itemsep1pt
\item \textbf{Identity.} If $R=I$ the numerator vanishes and $\mathrm{Pay}=1$ exactly. This is the
      invariant that caught three earlier fitted versions of the guard, none of which reached it.
\item \textbf{Total loss.} If $R$ maps every state to one point, $P(R\phi)$ is constant, the
      numerator becomes $\sum\lVert P(\phi)-P_0\rVert^2$ for a fixed $P_0$, which is at least the
      denominator, so the ratio is $\ge1$ and the score is at most $0$ --- reaching the $-1$ clip
      when $P_0$ is far from $\bar P$. Negative values therefore mean \emph{worse than predicting
      the ensemble mean}, and the clip exists so that distance is not mistaken for a measurement.
\item \textbf{Any invertible linear map of the probes.} For $P\mapsto AP$ with $A$ invertible, the
      covariance becomes $A\Sigma A^\top$, so each Mahalanobis term $v^\top\Sigma^{-1}v$ is
      unchanged and $\mathrm{Pay}$ is unchanged with it. This covers scalar rescaling, unequal
      per-component rescaling $P\mapsto\Lambda P$, and any change of basis. Translation
      $P\mapsto P+b$ is likewise absorbed, since the numerator is a difference and the denominator
      is measured about the mean. The score is therefore a property of the reduction and the probe
      \emph{span}, not of the units or coordinates the probes are written in.

      The unweighted version this replaced had none of that: it was invariant under a scalar $c$
      only, so an observable with a larger numerical scale dominated the sum, and a threshold could
      not be carried between systems whose probes had different units. Destroying a large-scale
      component then scored $0.00$ while destroying a small-scale one scored $1.00$ for the same
      amount of damage; weighted they score $0.66$ and $0.67$. \texttt{verify.py} asserts both this
      and $\mathrm{Pay}(I)=1$.
\end{enumerate}

\paragraph{Why the pay guard is not optional under a parsimony objective.} The constant reduction
is the most parsimonious candidate there is: one constant variable and the mechanism $\dot q=0$.
An objective that scores a coarse description by its simplicity and by how well it predicts its
own future therefore ranks it above every real reduction, since it is the global minimiser of
both terms. The mechanistic world-model objective of \citet{posner2026mwm}, their Eq.~9, combines
a complexity penalty on variables, mechanisms and types, a reuse term on the binding structures
across environments, and a prediction loss whose target Eq.~9 leaves unstated; their Sec.~2 describes prediction as
inferring future observations. The objective has exactly
that form when the prediction loss is taken in the learned coordinates rather than on decoded
observations, and the reuse term does not change the ranking, since one mechanism serving every
environment is maximal reuse. They ask
that variables minimise complexity while maximising explanatory scope. Closure scores neither
half; the pay guard of Eq.~\ref{eq:pay} scores the second, and App.~\ref{app:external} measures
what closure accepts on \dystsN\ systems without it.

\paragraph{Charge loss is reported, never blocked.} Discarding state detail can destroy a conserved
quantity while still closing. Physics does this deliberately, so the mechanism states which charges
a reduction loses rather than refusing it; on the lattice reduction both flavour charges are lost
(drift $7.5{\times}10^{-1}$ and $1.9{\times}10^{0}$), and it says so.

\paragraph{L3, transfer.} Transfer finds a map carrying one domain's states into another's, so that
what the first knows becomes a claim about the second. The structural guard requires a fitted linear map to make the target's
probes agree with the source's to within 5\%, and is scored on states the map was \emph{not} fitted
to. The second guard pushes the source's conserved charge through that map and requires the
target's own evolution to conserve it to $10^{-3}$. Table~\ref{tab:l3} gives six seeds, each with
a matched control whose symmetry is deliberately broken.

\begin{table}[h]
\caption{L3 transfer across six seeds, each with a matched control whose symmetry is deliberately
broken. The structure test is scored both on the 40 states the map was fitted to and on 40 disjoint
states; only the held-out column is used to accept. Scoring it in-sample, as an earlier version of
this work did, overstates its failure rate.}
\label{tab:l3}
\centering\footnotesize
\begin{tabular}{@{}lL{2.1cm}L{3.6cm}L{4.4cm}@{}}
\toprule
 & structure, in-sample & structure, \textbf{held out} ($<0.05$) & charge test ($<10^{-3}$) \\
\midrule
real    & median $0.0057$ --- 6/6 & median $0.0061$, worst $0.0463$ --- \textbf{6/6 pass} & median $1.3{\times}10^{-11}$, worst $5.9{\times}10^{-11}$ --- \textbf{6/6 pass} \\
control & median $0.0455$ --- 5/6 & median $0.0520$, best $0.0405$ --- \textbf{2/6 pass} & median $5.2$, best $2.8$ --- \textbf{0/6 pass} \\
\bottomrule
\end{tabular}
\end{table}

\paragraph{L4, positing an entity.} L4 consumes what L1's symmetry branch produces: when a found
charge fails to balance across an observed process, the missing amount describes something not yet
named. Writing $\Delta Q$ for charge in minus charge out, the
operation posits a carrier of $\Delta Q$ when $\Delta Q \neq 0$ and is silent otherwise. The second
guard requires $|\Delta Q|$ to exceed the \emph{noise scale}, not merely to sit many standard
errors from zero. It fires only when the missing charge exceeds three times the 2\% measurement
noise, that is $\LfourThresholdPct\%$ of the input charge. Its nulls have nothing hidden and the
measured output miscalibrated by $\LfourOffsetA\%$ and $\LfourOffsetB\%$.

\section{Results and controls in full}
\label{app:results}

The four operations have so far appeared only as things that fail in a particular way. They also
work, and NAR is only interpretable if they do --- a guard pair that accepted nothing would score
$\mathrm{NAR}=0$ while being useless, which is the reading
Table~\ref{tab:positives} was built to exclude. This section is what the operations recover when
both guards are applied. Negative results are reported as found, and twenty-one engineering checks
on cases whose answers we supplied all pass.

\paragraph{L1: recovering and classifying a hidden symmetry from a supplied generator basis.} On a periodic $6{\times}6$ lattice carrying a
complex two-flavour matter field and U(1) links, probed only by gauge-invariant, flavour-blind
quantities, the mechanism is handed eight candidate generators and told nothing about which are
real. It \textbf{finds five invariant directions} and, by the locality test, splits them into
\textbf{one redundancy} and \textbf{four symmetries}, emitting four conserved charges with drift
$1.8{\times}10^{-11}$ to $7.6{\times}10^{-10}$ against the redundancy's would-be charge at
$8.6{\times}10^{-1}$ --- nine orders of magnitude, at the trajectory's own precision. At uniform
order the gauge transformation and the global phase act identically; only the local test separates
them, and without it all four charges are destroyed to remove one redundancy.

\paragraph{L2, L3, L4.} The reduction is accepted at $62.9\times$ and $25.2\times$ separation and
refused below, placing the floor between $12\times$ and $25\times$. On the lattice, keeping 6 of 72
modes, it accepts \emph{nothing}, with learned and derived closure agreeing to within $0.002$
($1.628$ against $1.626$), so that negative is the physics rather than our fitting. Transfer accepts real
cases $6/6$ and matched controls $0/6$. Positing recovers a hidden carrier of $+1.192$ per event
against a withheld truth of $1.200$. It is silent when the books balance, when the imbalance is
noise, and when the measured output is miscalibrated by a few percent.

Appendix~\ref{app:silent} reports what happens when L1 is applied to a live
astrophysical controversy with no known answer: the criterion is \emph{silent}, because it needs
two descriptions that agree on the data and these disagree by ordinary model comparison.

\paragraph{Engineering checks (21/21).} Every case has an answer we supplied, so passing shows the
code does what it claims. The mechanism finds the unobservable direction in a description whose
concepts are never separable; quotients it away with predictions unchanged to $10^{-6}$;
\textbf{declines to act} on an honest theory with nothing hidden; \textbf{rejects} a patch whose new
concept moves no prediction, which is the property that stops ``add a parameter'' from being free;
and accepts a new concept only when the residual really has structure the old language lacked.

\paragraph{L1 in full.} The uniform Jacobian has rank 3 of 8, so five directions are invariant; a
link phase, a rescale and a single-flavour shift are rejected. The local test then isolates one
redundancy (local response $2.6{\times}10^{-5}$) from four symmetries
($5.1$--$6.6{\times}10^{1}$).

\paragraph{L2 and L4 sensitivity.} A mechanism that accepted at every separation would be broken;
this one has a measurable floor between $12\times$ and $25\times$. For L4, a real unseen product
carrying $\LfourHiddenA\%$ or $\LfourHiddenB\%$ of the charge gives an imbalance of
$\LfourHiddenZa\sigma$ or $\LfourHiddenZb\sigma$, and the operation stays silent because both sit
below its $\LfourThresholdPct\%$ threshold. This is the cost of the threshold: a carrier that small
is missed. The energy deficit in beta decay, the case the operation is modelled on, was far larger
than a few percent.

\paragraph{The proposer is the limiting component.} Across three runs, a world
whose residual hides $0.4\sin 3x$ yielded eight proposals of which two to five were accepted; a
control whose residual is pure noise yielded eight proposals of which \textbf{none} was accepted,
in every run. Offered the true concept directly the filter separates cleanly: $\sin 3x$ reaches 97\% coverage
against 68\% for $\sin 2.9x$ and 20\% for wrong terms. In the world hiding
$\sin 3x$ the model offered $\sin \pi x$, $\sin 2\pi x$, $\sin 3.7x$ and related terms, never
$\sin 3x$.

\subsection{Where the criterion goes silent, on real data}
\label{app:silent}

We applied L1 to a live controversy with
no known answer: the rotation curve of NGC 3198 from SPARC~\citep{lelli2016sparc}, fitted by a
dark-matter halo and by MOND~\citep{milgrom1983mond} with its acceleration scale fixed and free.
All three descriptions carry \emph{zero}
unobservable directions --- the halo model's softest direction still moves the curve $6.4\sigma$
under a 100\% excursion, unsurprising since SPARC's photometry was assembled to break exactly that
degeneracy. \textbf{The criterion is silent.} Its signature needs two
descriptions that \emph{agree} on the data; these disagree at $\chi^2/N$ of $1.22$ against $9.21$,
where ordinary model comparison settles it. The operation applies to a narrower class of disputes
than we assumed.

\section{The audit protocol, stated}
\label{app:formal}

Eq.~\ref{eq:nar} and Eq.~\ref{eq:car} define what we report. Four things need saying about them.

\paragraph{These are estimates against a list, not worst cases.} An earlier version of this paper
defined the quantity as a supremum, $\sup_{c\in\mathcal V}\Pr[G(c;D)=1]$, while reporting fractions
of controls accepted. Those disagree: for a deterministic guard the supremum is $1$ as soon as any
null passes. NAR is the mean we were computing all along and AnyNullPass is what the supremum was
reaching for, so both are now reported and neither is called a worst case. We do not optimise over
a candidate family, so no supremum claim would be warranted.

\paragraph{$V$ must be enumerated and typed --- and we chose it.} A guard with $\mathrm{NAR}=0$
against a weak $V$ has been told nothing, so we list the suite and label each member by what it
preserves and destroys (Table~\ref{tab:controltypes}).

\begin{table}[h]
\caption{The null suite is typed rather than pooled: a guard that survives a random direction but not a structure-preserving one has told us something specific, and pooling them into one rate hides it.}
\label{tab:controltypes}
\centering\small

\begin{tabular}{@{}lll@{}}
\toprule
type & preserves & destroys \\
\midrule
\textsc{degenerate}              & nothing & all structure (constant reduction, zero map) \\
\textsc{randomised}              & dimensionality & any relation to the source \\
\textsc{correspondence-breaking} & marginal statistics & the pairing between source and target \\
\textsc{structure-preserving}    & the invariant itself & only its redundancy status \\
\textsc{task-confounded}         & a trained model's geometry & the semantic content of the input \\
\textsc{sub-threshold}           & an effect's significance & its size against the noise scale \\
\bottomrule
\end{tabular}
\end{table}

\textbf{Choices made after a first run.} X1's first three nulls gave guard~1 an NAR of $0.00$. We
judged them too weak and added stronger ones. The current suite (Table~\ref{tab:x1cands}) has since
replaced all of them with correspondence-breaking nulls that share the vocabulary
(Appendix~\ref{app:xai}). Stronger nulls raise guard~1's NAR and make guard~2's test harder. X2's
selection rule changed after a first attempt returned features whose top tokens occurred 3 to 29
times, and its analysis changed after its first matched-control run (Appendix~\ref{app:xai}).
Per-control results are reported throughout, so a reader can recompute NAR over any sub-suite.

\paragraph{How diverse is a null suite?} Counting nulls overstates a suite whose mass sits on one
construction: six nulls that are five variants of one degenerate construction are not six
independent ways of being empty. We therefore report the effective number of types a suite
exercises, $\exp$ of the entropy of its type mix \citep{jost2006entropy}, which falls toward $1$ as
one type dominates and reaches the nominal count only when types are balanced. Our suites give
$\XoneNullEffTypes$ of $\XoneNullTypes$ types for X1, $\XtwoNullEffTypes$ of $\XtwoNullTypes$ for
X2, and $\XthreeNullEffTypes$ of $\XthreeNullTypes$ for X3, so no headline NAR here is carried by a
single construction. \citet{song2026shape} use the same statistic for the reverse purpose, to measure how an LLM spreads
reasoning effort over interpretations of a problem.

\paragraph{The vacuous repair, as a null type.} Our null types are degenerate, randomised,
correspondence-breaking, structure-preserving, task-confounded and sub-threshold. \citet{zheng2026far} name a
specific member of the structure-preserving family that we had not, and that generalises well
beyond their setting: the \emph{vacuous repair}. When a candidate fails, a discovery system may
weaken it until it is trivially true rather than abandon it, and the weakened statement keeps the
objects, the notation and the shape of the original while carrying none of its content. Their
pipeline guards against this by instructing the prover to check that a repair is not ``so weakened
that it no longer captures the intended conjecture''.

This is the same failure as the constant reduction of Sec.~\ref{sec:guard}, in a different domain.
A reduction mapping every state to zero is a coarse-graining weakened until closure is trivial; a
vacuously repaired conjecture is a statement weakened until proof is trivial. Both keep the form
and lose the content, and both are accepted by a guard that scores form. The recipe transfers: for
any operation that can weaken a candidate rather than reject it, the weakened-to-triviality version
belongs in $V$, and its acceptance rate is the quantity to report. We recommend it as a default
member of the null suite for systems that repair candidates, and note that we have not run it here,
since none of our four operations repairs a failed candidate.

\paragraph{A descriptive instrument that states its own gap.} SHAPE labels chain-of-thought traces
with mathematical heuristics and tracks the \emph{semantic space} the model is working in, then
reports how reasoning effort is distributed over those spaces. Its heuristic tagger is validated
against an adjudicated reference set. Its semantic-space tracker is not, and the authors say why:
the spaces are trajectory-relative interpretations for which, in their words, no objective gold
standard exists, so the tracker is calibrated by manual review instead. By the distinction drawn below (the two kinds of guard) it
is a structural score, computed on the trajectory it is describing and committing to nothing that
could fail, and the paper's structural findings rest on it. What would test it is a null suite: trajectories carrying no coherent
interpretation, on which a tracker that labels whatever it is given would still return a well-formed
sequence of spaces. \textbf{We ran that audit and it was inconclusive}, and we report what running it
required. Using the authors' own coding
manuals and metric code, on \ShapeProblems\ MATH-500 problems, we compared a model's own trace
against three typed nulls (its content units reordered, a trace for a different problem, units
spliced from unrelated traces) and one positive control (two independent solutions to the same
problem concatenated, which contains a genuine restart). Our annotator is a 7B model, not the 27B
one they use, because that model does not fit the hardware here. At that scale the tracker takes
\ShapeNew\ \textsc{new} decisions and \ShapeReturn\ \textsc{return} decisions across
\ShapeTraj\ trajectories. Every condition sits between $\ShapeEffLo$ and $\ShapeEffHi$ effective
spaces, and nothing separates: the positive control itself reaches AUROC $\ShapePosAUROC$. That
is a floor effect in our annotator, not a finding about their tracker, and the null AUROCs of
$\ShapeNullLo$ to $\ShapeNullHi$ carry no information. The audit is shipped so that anyone with the
27B annotator can run it in one command.

\paragraph{The two kinds of guard.} A \textbf{structural score} $S(c,D)$ is computed on the data
the candidate was fitted to; a \textbf{consequence test} $H_c(D_{\text{test}})$ requires the
candidate to commit in advance to an outcome on information not used to fit it. The evidence here
supports a conditional claim rather than a general one, and the condition is not the one we first
proposed. We expected $S$-guards to have high NAR only \emph{when the candidate has freedom
relative to $D$}, so that more data would fix them. That holds for the fitted map: scored in-sample, $R^2$ accepts $\XoneNARinsample$ of X1's nulls at
eight stimuli and $\XoneNARinsampleLarge$ at $\XoneSweepMax$. It does not explain the held-out figure of $\XoneNARstructural$, or
$\XthreeNARstructural$ for the protein models. The residue is a second mechanism entirely: linear CKA is a function of the two sets of
representations, so it is invariant to \emph{every} claim about how they correspond, and no amount
of data touches that. X2's co-occurrence count has little freedom and no such invariance, and
reaches $\XtwoNAR$ against a weak suite (Appendix~\ref{app:xai}). Consequence tests fail in other ways, and reporting both kinds of guard
separates the two.

\begin{table}[h]
\caption{The invariants that caught our own wrong guards. A version violating one of these is wrong however plausible its numbers look; the first two found three earlier pay guards and an early causal guard, and the third found
L3's in-sample structural score.}
\label{tab:invariants}
\centering\small

\begin{tabular}{@{}ll@{}}
\toprule
a reduction that keeps everything & must score exactly $1.000$ \\
a guard & must never rank a null candidate above the real one \\
a structural score & must be reported on data the candidate was not fitted to \\
\bottomrule
\end{tabular}
\end{table}

\begin{table}[h]
\caption{The two failure modes of Sec.~\ref{sec:framework}. The physics operations and X1 and X3
show the first; X2 shows the second.}
\label{tab:modes}
\centering\small
\begin{tabular}{@{}lll@{}}
\toprule
 & the guard accepts\ldots & seen in \\
\midrule
null acceptance (vacuity) & candidates carrying nothing the operation seeks & L1--L4, X1, X3 \\
unsupported acceptance & candidates an independent test does not support & X2 ($\XtwoFeatures$ paired features) \\
\bottomrule
\end{tabular}
\end{table}

\paragraph{Low NAR is necessary, not sufficient.} X2 is the counterexample: a guard rejecting
every null still leaves $\XtwoUnsupportedPct$ of what it accepts unsupported by the ablation test
(Table~\ref{tab:x2car}), and $\XtwoNeither$ unsupported by either of two interventions. A complete report of a guard needs NAR against a typed suite \emph{and}
CAR against a consequence test fixed before it is scored.

\section{A language model on the reinterpretation task}
\label{sec:llm}

A language model can recount the Lorentz--Einstein episode, so it is fair to ask whether the
operation is already available in a system that has read the histories. We put five episodes, three historical and two constructed controls, to a 20B open-weight
model~\citep{openai2025gptoss} in three conditions: \textbf{named}; \textbf{blind}, every identifier alpha-renamed with fit
statistics retained; and \textbf{bare}, renamed with statistics withheld. Table~\ref{tab:llmfull}
breaks the result out by episode.

\begin{table}[h]
\caption{Accuracy at identifying which of two candidates is the reinterpretation. Twelve repeats
per cell, presentation order randomised, chance 50\%.}
\label{tab:llmfull}
\centering\footnotesize
\begin{tabular}{@{}p{7.9cm}ccc@{}}
\toprule
episode & named & blind & bare \\
\midrule
dark energy~\citep{riess1998acceleration,perlmutter1999measurements} vs.\ grey dust~\citep{aguirre1999dust} & 67\% & 42\% & 50\% \\
relativistic mass vs.\ Lorentz's aether~\citep{lorentz1904electromagnetic,einstein1905electrodynamics} & 100\% & 100\% & 100\% \\
quasicrystals~\citep{shechtman1984quasicrystal} vs.\ twinning~\citep{pauling1985twinning} & 100\% & 50\% & 42\% \\
two constructed controls & 83\% & 25\% & 38\% \\
\midrule
pooled, 60 trials & \textbf{87\%} & 48\% & 53\% \\
\bottomrule
\end{tabular}
\end{table}

Accuracy is far above chance when the model can recognise the episode ($p=5{\times}10^{-9}$) and
\emph{indistinguishable from chance} when it cannot ($p=0.9$ blind, $0.7$ bare; named against
blind, paired $p=0.03$). On these five episodes it behaves as though recalling rather than judging. We do not claim this
generalises beyond them, and renaming alone is already known to move accuracy on grade-school arithmetic~\citep{mirzadeh2025gsmsymbolic}. One case
survives anonymisation: relativity stays at 100\% because $(1-v^2)^{-1/2}$ is recognisable from its
algebra whatever the identifiers are called. Withholding or supplying the statistics rescues
neither, so this is not a missing-information problem.

\section{What probe-based discovery cannot reach}
\label{sec:bound}

The machinery above buys concepts with measurements. Tested against eight concept-forming episodes
in the Standard Model it accounts for \textbf{two}: the neutrino, forced by a continuous spectrum
where a two-body decay demands a line, and gauge redundancy. \textbf{Four of the other six contain
no probe at all.} Colour was invented so the $\Delta^{++}$ state would be \emph{allowed} under
Fermi statistics; charm so that flavour-changing neutral currents would \emph{cancel}; complete
generations so that gauge anomalies would sum to zero; and the Higgs below about a TeV so that
$WW$ scattering would stay unitary --- which told experimenters what to build.

Each predicted a real particle, and each was justified by a \textbf{theorem first}, with the
experiment arriving years or decades later. A framework that pays for a concept with an experiment
does not reach these. Reaching them needs a formal theory in which to prove theorems, and no
discovery system we know of carries one. We see this as a main limit on automated physics
discovery today.

\section{The three interpretability experiments in full}
\label{app:xai}

\begin{table}[h]
\centering
\caption{X1 and X3, medians across configurations. Guard~1 accepts when $R^2$ or CKA exceeds
$0.5$. Causal: median correlation between the two models' induced shifts; guard~2 accepts above
$\XoneCausalMin$. AUROC: separation of a candidate's per-intervention scores from the genuine
pair's. Accepted: both guards accept. \textbf{All nulls except the randomly initialised
target, and also the donor-permuted control, have the same CKA as the genuine correspondence} to three
decimals, because CKA
scores the two sets of representations and never looks at the claimed map. Rows are generated from
the result files.}
\label{tab:x1cands}
\footnotesize
\begin{tabular}{@{}llrrcrrc@{}}
\toprule
candidate & type & $R^2$ & CKA & guard 1 & causal & AUROC & accepted \\
\midrule
\multicolumn{8}{@{}l@{}}{\emph{X1: Pythia pairs, $\XoneConfigs$ configurations, $\XoneTriples$ interventions each}} \\
the correspondence under test & real & 0.82 & 0.61 & 4/4 & +0.80 & -- & 4/4 \\
pairings permuted & corr.-breaking & -0.04 & 0.61 & 3/4 & -0.00 & 0.995 & 0/4 \\
the same map, randomly rotated & corr.-breaking & -0.96 & 0.61 & 3/4 & -0.01 & 0.996 & 0/4 \\
covariance-matched Gaussian target & randomised & -0.04 & 0.61 & 3/4 & +0.00 & 0.995 & 0/4 \\
fitted on disjoint text & task-confounded & 0.65 & 0.61 & 3/4 & +0.01 & 0.996 & 0/4 \\
randomly initialised target & degenerate & 0.19 & 0.03 & 0/4 & -0.00 & 1.000 & 0/4 \\
donor permuted at patch time & test control & 0.82 & 0.61 & 4/4 & -0.09 & 0.965 & 0/4
 \\
\midrule
\multicolumn{8}{@{}l@{}}{\emph{X3: ESM-2 pairs, $\XthreeConfigs$ configurations, $\XthreeTriples$ interventions each}} \\
identity map (code check) & positive & 1.00 & 1.00 & 3/3 & +1.00 & -- & 3/3 \\
adjacent layer, same model & positive & 0.73 & 0.76 & 3/3 & +1.00 & -- & 3/3 \\
the correspondence under test & real & 0.47 & 0.65 & 3/3 & +1.00 & -- & 3/3 \\
pairings permuted & corr.-breaking & -0.02 & 0.65 & 3/3 & +0.01 & 1.000 & 0/3 \\
residues permuted within protein & corr.-breaking & 0.12 & 0.65 & 3/3 & +0.02 & 1.000 & 0/3 \\
the same map, randomly rotated & corr.-breaking & -0.64 & 0.65 & 3/3 & +0.01 & 1.000 & 0/3 \\
covariance-matched Gaussian target & randomised & -0.02 & 0.65 & 3/3 & +0.01 & 1.000 & 0/3 \\
randomly initialised target & degenerate & 0.39 & 0.21 & 0/3 & +0.03 & 1.000 & 0/3 \\
donor permuted at patch time & test control & 0.47 & 0.65 & 3/3 & -0.16 & 0.974 & 0/3
 \\
\bottomrule
\end{tabular}
\end{table}

X1 and X2 use models from the Pythia suite~\citep{biderman2023pythia}, whose members share a
tokenizer, which is what makes the causal comparison in X1 possible at all; their text is
WikiText-2. X3 uses ESM-2. Everything is scored on held-out data.

\paragraph{X1 setup.} $\XoneConfigs$ configurations over the Pythia suite: 70m~L4$\to$160m~L8,
70m~L2$\to$160m~L4, 160m~L8$\to$410m~L12 and 70m~L4$\to$410m~L12. Text is 240 WikiText-2
sequences of 128 tokens; the map is fitted on half and everything is scored on the other half. Guard~1 is ridge regression
($\lambda=10^{2}$) reporting held-out $R^2$ about the \emph{fitting} mean, plus linear CKA,
accepting at $R^2>0.5$ or $\mathrm{CKA}>0.5$. Guard~2 runs $\XoneTriples$ interventions per
configuration ($\XoneSeeds$ seeds $\times$ $\XonePos$ positions $\times$ $\XoneDonors$ donors, each
scored over $\XoneRecip$ recipient sequences). Each adds the mapped difference at a single
position and correlates the two models' induced shifts after centring across donors. Guard~2
accepts when the median correlation exceeds $\XoneCausalMin$, and a candidate is accepted when both
guards accept.

\paragraph{Each candidate is scored with the map it claims.} A candidate that composes the fitted
map with a rotation is a claim about \emph{that} composed map, and an earlier version scored every
candidate with the real map, so the rotated and donor-permuted rows reported $R^2$ identical to the
genuine pair. Scored correctly the rotation falls to $-0.96$, while its CKA does not change.

\paragraph{An earlier pathology, and what removed it.} We previously reported that our own causal
guard preferred a vacuous candidate: correlating \emph{uncentred} shifts ranked a random-token
control at $0.70$ against $0.39$ for the genuine pair, and centring across donors was what fixed
it. That was an artefact of the intervention rather than of the metric. The old guard overwrote the
residual stream at every position with one mean-pooled donor vector, which pushed both models
toward the same frequent tokens, and that shared generic response dominated the correlation. Under
localised difference patching the generic response is small: the worst null now scores $\XoneWorstNullRaw$
uncentred against $\XoneWeakestRealRaw$ for the weakest genuine pair. Centring is still applied, and is no longer
load-bearing. We report this because the earlier claim appeared in a previous version of this
work.

The earlier null suite also included ChemBERTa~\citep{chithrananda2020chemberta} reading SMILES as
a different-domain control, for which guard~2 was \emph{undefined}: the causal test compares two
next-token distributions and a chemistry model shares no vocabulary with an English one. The
rebuilt suite replaces it with correspondence-breaking nulls that share the vocabulary, which is a
harder test, and the limitation it exposed
stands: the refutable guard needs a shared output space to compare in.

\paragraph{X3 setup.} $\XthreeConfigs$ ESM-2~\citep{lin2023esm2} scale pairs (8M~L3$\to$35M~L6,
8M~L4$\to$150M~L15 and 35M~L6$\to$150M~L15) on $\XthreeProteins$ UniRef50 sequences truncated to
$\XthreeResidues$ residues, half fitted and half held out, $\XthreeTriples$ interventions per
configuration. Guards are \emph{imported} from the X1 module, not reimplemented. Two candidates are
positive controls rather than nulls: an identity map at the same layer, which cannot fail unless
the code is broken and is there to check exactly that, and the adjacent layer of the same model,
which is a genuine correspondence the guard must find. Both are accepted in every configuration.

\paragraph{X2 setup.} One TopK sparse autoencoder per arm, on Pythia-160m layers 4 and 8 and
Pythia-410m-deduped layers 8 and 16, $8\times$ expansion (6{,}144 or 8{,}192 features), $L_0=32$, recovering 93--99\% of activation variance depending on the arm;
1{,}200 sequences split 60/20/20 into autoencoder training, guard scoring and the consequence test,
so a feature is selected on one split and judged on another. An $L_1$ penalty was tried first and would not sparsify --- at the coefficient that left
reconstruction usable, $L_0$ settled near 565 and the most frequently firing features were generic
ones whose top token was simply the corpus's most common token. TopK sets $L_0$ by construction.

\paragraph{Two measurement pitfalls we hit, both of which inflate guard 1.} First, ranking features
by raw firing frequency selects the always-on ones; every "most interpretable" feature it returned
had \emph{the} as its top token. Second, and less obvious: scoring a feature by the likelihood
ratio $P(\text{token}\mid\text{fires})/P(\text{token})$ is degenerate when the feature captures
every occurrence of a rare token, because the ratio then collapses to $N/(\text{times fired})$ and
is independent of how many occurrences there were. The twelve "most selective" features it first
returned had top tokens occurring 3 to 29 times in held-out text --- too few to intervene on. We
therefore require a candidate concept to occur at least 60 times in the guard split before it is
scored.

\paragraph{X2 nulls.} In each arm guard~1 scores four nulls. The first is a random unit direction
in the residual space. The second is the decoder vector of the arm's most enriched feature with its
coordinates permuted, and the third is the same vector, unpermuted, plus Gaussian noise of $0.6$
times its norm. The fourth is the feature with the lowest nonzero firing rate. The first three are read as the residual stream's projection
on the direction, clipped at zero. All sixteen score enrichment of at most $\XtwoNullEnrHi$ against
the threshold of 4. The surrogate keeps little of its source feature's enrichment
($\XtwoSurrLo$--$\XtwoSurrHi$ against $\XtwoTopEnrLo$--$\XtwoTopEnrHi$), so this suite tests
little. In the first run to the specification, under the earlier concept definition below, the
surrogate passed guard~1 in one arm (enrichment $\XtwoSpecNullPassEnr$, $\mathrm{NAR}=\XtwoSpecNAR$).

\paragraph{Full X2 result.} Table~\ref{tab:x2flow} gives the selection flow. A feature is
selected at enrichment $>4$; support requires the standardised difference
$S_f=(\Delta L_C-\Delta L_{\neg C})/\mathrm{SE}[\cdot]$ to exceed $1.96$ with $\Delta L_C>0$. An
earlier version thresholded the raw ratio $\Delta L_C/\Delta L_{\neg C}$, which is unstable when the denominator is near zero, where one feature scored $1644$. The
standardised difference replaces it. This is stricter, and the reported CAR is correspondingly lower than the ratio gave.

\begin{table}[h]
\caption{X2 selection flow. Features are ranked by enrichment and the top 150 per arm are taken, so
the fact that every selected feature clears the threshold is a property of the pipeline and is not
reported as a result. Between $\XtwoFlowLo$ and $\XtwoFlowHi$ features per arm clear it; we test $\XtwoNfeat$. Rows are generated from the result file.}
\label{tab:x2flow}
\centering\small
\begin{tabular}{@{}lrrrr@{}}
\toprule
stage & 160m L4 & 160m L8 & 410m L16 & 410m L8 \\
\midrule
all SAE features & 6144 & 6144 & 8192 & 8192 \\
firing rate in range & 3772 & 1248 & 2515 & 3540 \\
concept frequent enough & 3772 & 1248 & 2508 & 3540 \\
enrichment $>4$ & 1263 & 605 & 1169 & 1453 \\
\midrule
selected for intervention & 150 & 150 & 150 & 150
 \\
\bottomrule
\end{tabular}
\end{table}

\paragraph{The frozen specification and what changed after it.} \texttt{corpus/ANALYSIS.md} was
committed on 1 September 2026, before the first matched-control run. It fixes the endpoint
$\Delta\mathrm{CAR}$ and its retraction condition: if the interval includes zero, the X2 claim is
withdrawn. It also fixes the support statistic $S_f$, the selection rule and the matching rule. That run drew one
control per feature, matched on firing rate, and gave CAR $\XtwoSpecCAR$ against $\XtwoSpecControl$
over $\XtwoSpecN$ features: $\Delta\mathrm{CAR}=\XtwoSpecDelta$, positive in $\XtwoSpecArmsPositive$
of $\XtwoSpecArms$ arms. The pipeline was then revised, each change made after that result was
known.
(i) The data were split three ways, so that a feature's concept is chosen on one split and judged
on another.
(ii) A feature's concept became the next token; it had been the token at the firing position. This
changes both guard~1's concept and $H$'s mask.
(iii) The ablation became the subtraction of the feature's own contribution; it had erased the whole
decoder direction.
(iv) The standard error in $S_f$ became one over per-sequence contrasts; it had been over tokens.
(v) Each feature received three controls, also matched on mean activation, with the majority rule
of Sec.~\ref{sec:xai}. A control counts as supported when $S_f>1.96$. The first control alone gives
a control rate of $\XtwoControlFirst$ and $\Delta\mathrm{CAR}=\XtwoDeltaFirst$ (95\% CI
$\XtwoDeltaFirstCIlo$ to $\XtwoDeltaFirstCIhi$).
(vi) The interval resamples arms and concept clusters with each verdict held fixed; the
specification also resampled sequences, which counts the noise in $S_f$ twice.
(vii) Features too rare in the consequence split to be judged were removed from the denominator,
as the specification's exclusion rule requires. Counting them as unsupported gives CAR
$\XtwoCARallSelected$.
The specification states support as a bootstrap interval that excludes zero; both runs use
$S_f>1.96$. The positive control on $H$ is not in the specification. The first run's result file
ships as \texttt{xai\_sae\_spec\_run.json}.

\begin{table}[h]
\caption{The positive control on X2's consequence test, per arm. For each selected feature the
gradient of its concept token's logit, estimated on the consequence split, is ablated.
``Defined'': testable features for which the gradient could be computed (204 on the two 160m
arms, a different set from the 204 paired features of Table~\ref{tab:x2car}). ``Detected'': the rate at which $H$
supports the ablated direction ($S_f>1.96$). ``Wrong sign'': the rate at which ablating it damages prediction of
the concept \emph{less} than prediction elsewhere, so the control's own manipulation did not have its
intended effect. The control is used on the two arms where that rate is below $0.10$. Validity and detection use
the same score, so the detection rate on those arms is conditional on that choice. Rows are
generated from the result file.}
\label{tab:x2pos}
\centering\small
\begin{tabular}{@{}lrrr@{}}
\toprule
arm & defined & detected & wrong sign \\
\midrule
Pythia-160m L4 & 94 & 0.79 & 0.00 \\
Pythia-160m L8 & 110 & 1.00 & 0.00 \\
Pythia-410m L16 & 102 & 0.55 & 0.19 \\
Pythia-410m L8 & 99 & 0.07 & 0.93 
 \\
\bottomrule
\end{tabular}
\end{table}

The three splits are contiguous blocks of WikiText-2, so they hold different articles. Of the
$\XtwoUntestable$ untestable features, $\XtwoUntestableZero$ have no occurrence of their concept in
the consequence split and $\XtwoUntestableSome$ have at most $\XtwoUntestableMaxPos$ positions,
against the $\XtwoMinOn$ required; the smallest testable concept has $\XtwoTestableMinPos$ positions.
$\XtwoTopConceptN$ of the $\XtwoUntestable$ share one concept token, the ``send'' of ``Townsend'',
which occurs $\XtwoTopConceptGuard$ times in $\XtwoTopConceptGuardSeq$ sequences of the guard
split, all within one article, and $\XtwoTopConceptCons$ times in the consequence split. The
untestable features' median enrichment is $\XtwoEnrUntestable$ against $\XtwoEnrTestable$ for the
testable ones, so the exclusion does not select on the guard score.
CAR@$K$ at $K=\XtwoCurveK$ is $\XtwoCurve$ (Sec.~\ref{sec:framework}).

\paragraph{A second intervention.} Injecting the feature where it is silent and asking whether the
concept is helped is a different claim from removing what it contributes. Activation supports
$\XtwoCARactivate$ of the $\XtwoFeatures$ paired features: $\XtwoAgreeN$ of the $\XtwoAblationN$
that ablation supports, and $\XtwoActOnAblNeg$ of the $\XtwoAblNegN$ it does not. Part of the
disagreement is therefore specific to ablation: counting either intervention as support gives
$\XtwoEither$, and $\XtwoNeither$ of the paired features are supported by neither. Activation was
not the prespecified endpoint and has no matched controls, so it is reported beside CAR.

\paragraph{X1 and X3 per configuration.} Of the $\XoneNullsPerConfig$ nulls in each X1
configuration, in the order listed above, $R^2$ alone accepts $\XoneRtwoPerConfig$ and CKA alone
accepts $\XoneCkaPerConfig$. For X3 the counts are $\XthreeRtwoPerConfig$ and
$\XthreeCkaPerConfig$.

\paragraph{Reproduction.} \texttt{xai\_cka.py}, \texttt{xai\_esm.py} and \texttt{xai\_sae.py} in the release run the three
experiments end to end and write the tables above as JSON.

\section{Where the four operations came from}
\label{sec:corpus}
\label{app:corpus}
\label{app:prior}

\begin{figure}[t]
\centering
\includegraphics[width=\textwidth]{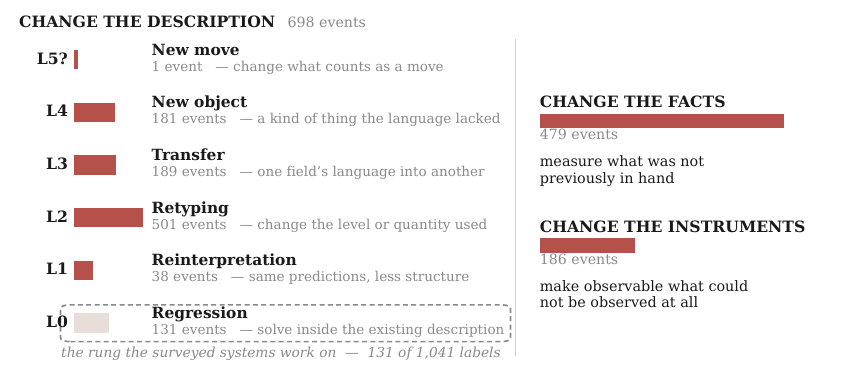}
\caption{What a discovering system has to be able to do. A move changes the facts you hold, the
instruments you have, or the description you use; only a change of description has a depth, and its
six rungs are \textbf{regression} (L0), solving inside the description you already have;
\textbf{reinterpretation} (L1), the same predictions with less structure; \textbf{retyping} (L2),
changing the level or quantity it is written in; \textbf{transfer} (L3), carrying another field's
language in; \textbf{new object} (L4), adding a kind of thing the language lacked; and \textbf{new
move} (L5), changing what counts as a move. The three column totals count annotated prize events
across all fields (1{,}363 of the 1{,}547; the other 184 fall in five smaller classes). The
per-rung counts are depth labels, since an event may sit on more than one rung (1{,}041 labels on
the 727 events that carry one). Bar lengths compare within a column: a square-root scale on the
ladder, a linear one on the right. The fourteen systems we surveyed operate within a predefined
representational or operator language; L5 is a single event its annotator flagged uncertain.}
\label{fig:ladder}
\end{figure}

The four operations were chosen from an annotation of \textbf{1{,}547 prize events} (2{,}327 award
records across 23 prizes, 1731--2026) by what the contribution did: it changed the \textbf{facts}
held, the \textbf{instruments} available, or the \textbf{description} used. Changes of description
form the ladder of Fig.~\ref{fig:ladder}. Across all fields 45\% of events change a description.
The physics subset (532 events) has nine physics prizes (the Nobel, Wolf, Dirac, Sakurai, Onsager,
Heineman, Boltzmann, Lorentz and Breakthrough Fundamental Physics prizes) plus the Kavli Prize,
whose fields include neuroscience and nanoscience. In it 65\% of events change a description, and
69\% with the Max Planck Medal in place of the Kavli Prize. This is why the operations are built for
physics. Sec.~\ref{app:ops} builds operations for four of the six rungs; L0 is what automated
discovery already does, and L5 was not attempted. The corpus has a single annotator, a language
model reading each official citation under human direction, and it supports no claim in this paper
(Sec.~\ref{sec:limits}).

Table~\ref{tab:prior} places fourteen automated-discovery systems on the ladder by the operation
each publishes, which is a statement about published operations and leaves open what each could be
extended to do. They cluster on L0; DreamCoder reaches L2 by naming reusable structure, and
predicate invention aims at L4. We found none that performs L1: none takes a theory that already
fits and returns one that fits identically while carrying less.

\begin{table}[t]
\caption{Fourteen automated-discovery systems placed on the ladder, by the operation each exposes. Most cluster on L0; \textbf{we found none that performs L1}.}
\label{tab:prior}
\centering\footnotesize
\begin{tabular}{@{}L{2.6cm}L{3.9cm}L{1.6cm}L{4.6cm}@{}}
\toprule
system & what it does & rung & why not higher \\
\midrule
BACON~\citep{langley1981bacon} & rediscovers laws from tables & L0 & searches a fixed term language \\
Eureqa~\citep{schmidt2009eureqa} & symbolic regression on motion & L0 & operator set given in advance \\
AI Feynman~\citep{udrescu2020feynman} & physics-informed symbolic regr. & L0 & recovers known forms; invents no term \\
PySR~\citep{cranmer2023pysr} & evolutionary symbolic regression & L0 & fixed primitive set \\
LaSR~\citep{grayeli2024lasr} & symbolic regr.\ with learned library & L0/L2 & abstractions are shortcuts, not retypings \\
DreamCoder~\citep{ellis2021dreamcoder} & wake--sleep library learning & \textbf{L2} & names structure; no theory to reinterpret \\
FunSearch~\citep{romeraparedes2024funsearch} & LLM program search + evaluator & L0 & searches a fixed language \\
AlphaGeometry~\citep{trinh2024alphageometry} & olympiad geometry & L0 & derives inside a fixed axiom set \\
Robot Scientist~\citep{king2004robot} & hypotheses with a robot & L0 $+$ facts & chooses experiments; ontology fixed \\
Coscientist~\citep{boiko2023coscientist} & autonomous chemistry & instruments & extends reach; no description changes \\
AI Scientist~\citep{lu2026aiscientist} & end-to-end ML papers & L0 & proposes and tests ideas; no description changes \\
AlphaFold~\citep{jumper2021alphafold} & protein structure prediction & tractability & makes computable; does not reframe \\
Cascade-Correlation~\citep{fahlman1990cascor} & grows hidden units as needed & L0 & capacity, not concepts \\
predicate invention~\citep{cropper2022ilp} & invents relations in logic programs & \textbf{L4}$^{*}$ & $^{*}$open after three decades of ILP \\
\bottomrule
\end{tabular}
\end{table}

\section{External replication: protocol details}
\label{app:external}

\begin{table}[t]
\centering
\caption{The L2 guards on a suite whose correct answer is fixed by construction. \textbf{Handed the
true subspace the guard accepts every one; handed the proposer's estimate of it, it accepts
$\posRecallEst$ of them.} The gap comes from the proposer. Meanwhile the closure test alone admits nearly two thirds of the empty
candidates, and the pair admits none.}
\label{tab:positives}
\small
\begin{tabular}{@{}lrrl@{}}
\toprule
candidate & $n$ & accepted by the pair & \\
\midrule
true subspace, known by construction & 24 & \textbf{24 / 24} & recall $1.00$ \\
the proposer's estimate of it        & 24 & \textbf{17 / 24} & recall $0.71$ \\
\midrule
fast, random, constant               & 72 & \textbf{0 / 72}  & precision $1.00$ \\
\quad \emph{closure test alone}      & 72 & \textbf{47 / 72} & $\mathrm{NAR}=0.65$ \\
\bottomrule
\end{tabular}
\end{table}

\begin{table}[t]
\centering
\caption{The L2 guards on \dystsN\ chaotic systems from \citet{gilpin2021chaos}, dimension 3--10,
with guards and thresholds imported unchanged and acceptance resting on the learned coarse law.
\textbf{The closure test accepts a reduction to nothing on \dystsConstPct\ of the benchmark. The pay
test refuses it on all of it.} The median closure error is over systems where it is finite. Rows are
generated from the result file.}
\label{tab:external}
\small
\begin{tabular}{@{}lrrrr@{}}
\toprule
reduction & closes & median closure err. & max pay score & \textbf{accepted by both} \\
\midrule
slow modes (the real candidate) & 0 / 129 & 1.810 & $+0.107$ & \textbf{0} \\ \midrule
\textsc{control} constant (all to zero) & 124 / 129 & 0.000 & $-0.041$ & \textbf{0} \\
\textsc{control} fast modes & 0 / 129 & 1.465 & $+0.065$ & \textbf{0} \\
\textsc{control} random subspace & 0 / 129 & 1.902 & $-0.012$ & \textbf{0}
 \\
\bottomrule
\end{tabular}
\end{table}


\paragraph{Tolerance.} On the labelled suite of Table~\ref{tab:positives}, whose systems are accepted
at every separation tested when untilted (a different construction from the system of
Table~\ref{tab:vacuity}, whose floor lies between $12\times$ and $25\times$), tilting the true subspace toward the fast one makes the
guard pair refuse it from
\posThetaBySep\ separation. At strong separation a small tilt already mixes in modes that are
much faster; at weak separation closure sits near its threshold before any tilt.

\paragraph{What was skipped and why.} Of 137 systems in \texttt{dysts}, 129 ran. Eight were
skipped: two abstract base classes with no concrete right-hand side, and six delay-differential
systems (\texttt{IkedaDelay}, \texttt{MackeyGlass}, \texttt{PiecewiseCircuit},
\texttt{ScrollDelay}, \texttt{SprottDelay}, \texttt{VossDelay}) whose state is a function on an
interval rather than a vector in $\mathbb{R}^d$, so a finite-dimensional linear reduction is not
defined for them. No system was dropped after seeing its result.

\paragraph{Jacobians.} The proposer needs a linear operator. Where \texttt{dysts} ships an analytic
Jacobian we use it; where it does not, we take central differences of the published right-hand
side, which is the same operator. Systems are not excluded for lacking an analytic Jacobian.

\paragraph{The five systems where the constant reduction does not close} are
\texttt{BlinkingRotlet}, \texttt{BlinkingVortex}, \texttt{InteriorSquirmer},
\texttt{SwingingAtwood} and \texttt{TurchinHanski}. These are numerical failures of evaluation, and
the closure condition itself holds for them. The constant reduction lifts its coarse state to the
fine-state origin, and there the published right-hand side of each is undefined: the first two
return NaN and the other three divide by zero. Both the derived and the learned coarse law then have
a non-finite right-hand side, the closure error is recorded as non-finite, and a non-finite error
counts as not closing. A law fitted to finite differences of the coarse trajectory would have zero
targets and would close. With the origin moved off the attractor all five close with error $0.000$,
and the zero-dimensional reduction, which evaluates no right-hand side, closes on all \dystsN. An
earlier version of this appendix attributed the five to time dependence and a non-invariant origin;
that explanation was wrong, since a constant coarse trajectory satisfies $\dot q=0$ whatever the
fine dynamics. The coordinate check runs on every system in \texttt{dysts}; the counts here are over the
\coordN\ of the main sweep, and it ships as \texttt{external\_coordinates.json}. Moving the origin makes the right-hand side of three other systems
(\texttt{WindmiReduced}, \texttt{YuWang}, \texttt{YuWang2}) non-finite at the displaced point, which
is why the lifted version closes on \coordTranslatedCloses\ there rather than on all.

\paragraph{The negative does not depend on how hard we compressed.} Keeping half the directions is
an aggressive cut, so we repeated the entire sweep at $k = d-1$, dropping a single direction, the
gentlest reduction that is still a reduction, with the same learned-only rule. The real candidate
then closes on \textbf{\mildSlowCloses\ of \mildN} systems and is accepted by both guards on
\mildSlowBoth; the constant reduction closes on \textbf{\mildConstCloses} and pays on
\textbf{\mildConstPays}; and \mildEmptyBoth\ of the $3\times\mildN$ empty controls survive both
guards. A linear subspace of the mean Jacobian is the wrong object for these curved attractors at
either cut. An earlier version of this paragraph reported this sweep under the superseded rule and
the unweighted pay score; the numbers here replace it.

\paragraph{One control behaves differently here than on our own system,} and the difference is
informative. On the slow--fast system of Table~\ref{tab:vacuity}, the \emph{fast}-mode reduction
closed perfectly because fast modes decay to zero. On these dissipative chaotic systems it does not close at all, because the fast
subspace of the mean Jacobian still carries attractor motion rather than decaying to nothing. The
vacuity is therefore carried entirely by the constant control here. Our claim narrows accordingly:
\emph{closure alone cannot distinguish a real reduction from a reduction to nothing}, which holds
on \dystsConstPct\ of an external benchmark, rather than the stronger claim that every empty reduction closes
in every setting.

\paragraph{Reproduction.} \texttt{corpus/mechanism/external\_dysts.py} in the release runs the
whole sweep, importing the guards from \texttt{l2\_coarse.py} beside it; its output ships as
\texttt{external\_dysts.json}. \texttt{fig\_external.py} regenerates Fig.~\ref{fig:external} from
that file and \emph{asserts} the two headline counts rather than trusting them, and the release's
\texttt{verify.py} re-checks both on every run.

\end{document}